\documentclass[10pt,twocolumn,letterpaper]{article}
\usepackage{authblk}
\usepackage[accsupp]{axessibility}

\usepackage{cvpr}              

\usepackage{graphicx}
\usepackage{amsmath}
\usepackage{amssymb}
\usepackage{booktabs}

\usepackage[accsupp]{axessibility}

\usepackage{bm}
\usepackage[export]{adjustbox}
\usepackage{amsfonts}
\usepackage{dsfont}
\usepackage[table]{xcolor}
\usepackage{soul}

\usepackage[pagebackref,breaklinks,colorlinks]{hyperref}

\usepackage[capitalize]{cleveref}
\crefname{section}{Sec.}{Secs.}
\Crefname{section}{Section}{Sections}
\Crefname{table}{Table}{Tables}
\crefname{table}{Tab.}{Tabs.}

\def\cvprPaperID{5934} 
\def\confName{CVPR}
\def\confYear{2023}

\begin{document}

\newcommand{\Tim}[1]{\textcolor{magenta}{#1}}
\newcommand{\Thomas}[1]{\textcolor{blue}{#1}}
\newcommand{\Behzad}[1]{\textcolor{red}{#1}}
\newcommand{\Tinne}[1]{\textcolor{green}{#1}}

\newcommand{\aug}{\tilde{\boldsymbol{x}}}
\newcommand{\im}{\boldsymbol{x}}
\newcommand{\glob}{\bar{\boldsymbol{z}}}
\newcommand{\dense}{\mathbf{Z}}
\newcommand{\pos}{\mathbf{E}}
\newcommand{\objects}{\mathbf{O}}
\newcommand{\centroids}{\mathbf{C}}
\newcommand{\costsem}{\mathbf{T}^{\text{(sem)}}}
\newcommand{\costpos}{\mathbf{T}^{\text{(pos)}}}
\newcommand{\costtot}{\mathbf{T}^{\text{(tot)}}}
\newcommand{\loc}{\boldsymbol{z}_{t,k}}
\newcommand{\p}{\boldsymbol{p}}
\newcommand{\PP}{\mathbf{P}}
\newcommand{\Q}{\mathbf{Q}}
\newcommand{\Y}{\mathbf{Y}}
\newcommand{\mask}{\mathbf{M}}
\newcommand{\tmarg}{\mathbf{r}}
\newcommand{\cmarg}{\mathbf{c}}
\newcommand{\cls}{\texttt{[CLS]}}
\newcommand{\model}{g}
\newcommand{\backbone}{f}
\newcommand{\head}{h}

\newcommand{\mname}{\text{CrOC}}
\newcommand{\vit}{\text{ViT}}
\newcommand{\dino}{\text{DINO}}
\newcommand{\mae}{\text{MAE}}
\newcommand{\ibot}{\text{iBOT}}
\newcommand{\beit}{\text{BEiT}}
\newcommand{\densecl}{\text{DenseCL}}
\newcommand{\resnet}{\text{ResNet}}
\newcommand{\soco}{\text{SoCo}}
\newcommand{\pixpro}{\text{PixPro}}
\newcommand{\resim}{\text{ReSim}}
\newcommand{\vicregl}{\text{VICRegL}}
\newcommand{\orl}{\text{ORL}}
\newcommand{\byol}{\text{BYOL}}
\newcommand{\detco}{\text{DetCo}}
\newcommand{\odin}{\text{ODIN}}
\newcommand{\cpsq}{\text{CP}^{2}}
\newcommand{\overbar}[1]{\mkern 1.5mu\overline{\mkern-1.5mu#1\mkern-1.5mu}\mkern 1.5mu}
\newcommand{\supp}{\textbf{Supplementary Material}}

\newcommand{\Raa}[1]{\textcolor{red}{#1}}
\newcommand{\Ra}{\Raa{rnBS} }
\newcommand{\Rbb}[1]{\textcolor{green}{#1}}
\newcommand{\Rb}{\Rbb{sP9Y} }
\newcommand{\Rcc}[1]{\textcolor{blue}{#1}}
\newcommand{\Rc}{\Rcc{97MY} }
\newcommand{\Rdd}[1]{\textcolor{orange}{#1}}
\newcommand{\Rd}{\Rdd{R4} }

\definecolor{light_cyan}{HTML}{c6effc}
\sethlcolor{light_cyan}

\title{CrOC \includegraphics[width=7mm]{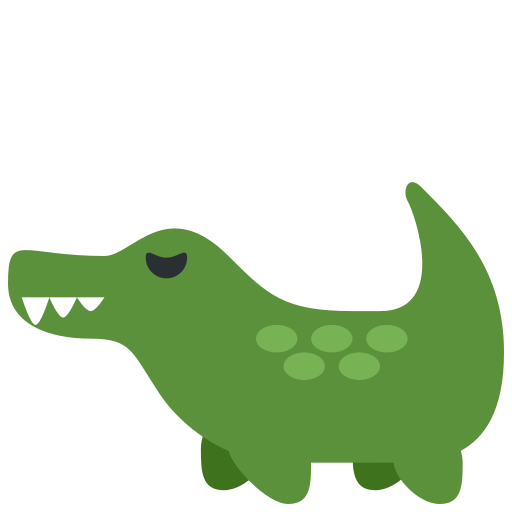}: Cross-View Online Clustering for Dense Visual Representation Learning}


\author{
Thomas Stegm\"uller$^{1*}$ \hspace{1cm} Tim Lebailly$^{2*}$ \hspace{1cm} Behzad Bozorgtabar$^{1,3}$ \\ \vspace{-0.35cm}
Tinne Tuytelaars$^{2}$ \hspace{1cm} Jean-Philippe Thiran$^{1,3}$ \\ \vspace{0.1cm}
$^{1}$EPFL \hspace{1cm} $^{2}$KU Leuven \hspace{1cm} $^{3}$CHUV \\ 
{\small $^{1}$\texttt{\{firstname\}.\{lastname\}@epfl.ch} \hspace{2cm} $^{2}$\texttt{\{firstname\}.\{lastname\}@esat.kuleuven.be}}
}


\maketitle

\begin{abstract}
Learning dense visual representations without labels is an arduous task and more so from scene-centric data. We propose to tackle this challenging problem by proposing a \textbf{Cr}oss-view consistency objective with an \textbf{O}nline \textbf{C}lustering mechanism (CrOC) to discover and segment the semantics of the views. In the absence of hand-crafted priors, the resulting method is more generalizable and does not require a cumbersome pre-processing step. More importantly, the clustering algorithm conjointly operates on the features of both views, thereby elegantly bypassing the issue of content not represented in both views and the ambiguous matching of objects from one crop to the other. We demonstrate excellent performance on linear and unsupervised segmentation transfer tasks on various datasets and similarly for video object segmentation. Our code and pre-trained models are publicly available at \textcolor{magenta}{https://github.com/stegmuel/CrOC}.
\end{abstract}
\newcommand\blfootnote[1]{%
  \begingroup
  \renewcommand\thefootnote{}\footnote{#1}%
  \addtocounter{footnote}{-1}%
  \endgroup
}
\blfootnote{* denotes equal contribution.}

\section{Introduction}
\label{sec:introduction}
Self-supervised learning (SSL) has gone a long and successful way since its beginning using carefully hand-crafted proxy tasks such as colorization \cite{larsson2017colorization}, jigsaw puzzle solving \cite{noroozi2016unsupervised}, or image rotations prediction \cite{gidaris2018unsupervised}. In recent years, a consensus seems to have been reached, and \textit{cross-view consistency} is used in almost all state-of-the-art (SOTA) visual SSL methods  \cite{chen2020simple,he2020momentum,grill2020bootstrap,caron2020unsupervised,caron2021emerging}. In that context, the whole training objective revolves around the consistency of representation in the presence of information-preserving transformations \cite{chen2020simple}, e.g., \textit{blurring}, \textit{cropping}, \textit{solarization}, etc. Although this approach is well grounded in learning image-level representations in the unrealistic scenario of \textit{object-centric} datasets, e.g., ImageNet \cite{5206848}, it cannot be trivially extended to accommodate \textit{scene-centric} datasets and even less to learn dense representations. Indeed, in the presence of complex scene images, the random \textit{cropping} operation used as image transformation loses its semantic-preserving property, as a single image can yield two crops bearing antipodean semantic content \cite{mo2021object,van2021revisiting,purushwalkam2020demystifying,selvaraju2021casting}. Along the same line, it's not clear how to relate sub-regions of the image from one crop to the other, which is necessary to derive a localized supervisory signal.
\par
To address the above issue, some methods \cite{mo2021object,selvaraju2021casting} constrain the location of the crops based on some heuristics and using a pre-processing step. This step is either not learnable or requires the use of a pre-trained model.
\begin{figure}
    \centering
    \includegraphics[width=\columnwidth]{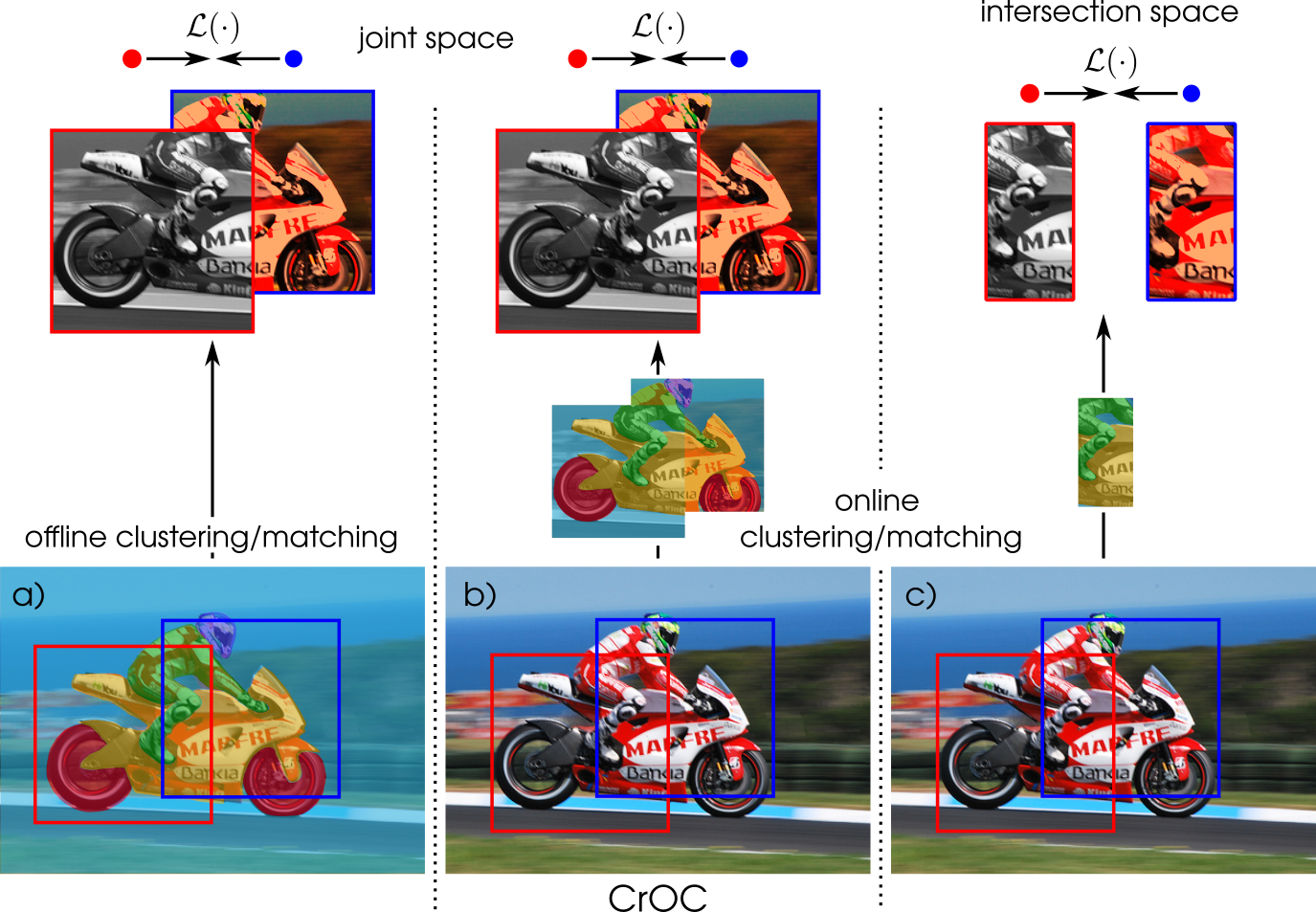}
    \caption{
    \textbf{Schematic for different categories of self-supervised learning methods for dense downstream tasks.} a) Prior to the training, a pre-trained model or color-based heuristic is used to produce the clustering/matching of the whole dataset. c) The matching/clustering is identified online but restrains the domain of application of the loss to the intersection of the two views. b) Our method takes the best of both worlds, leverages online clustering, and enforces constraints on the whole spatiality of the views.
    }
    \label{fig:joint_vs_intersect}
\end{figure}
Alternatively, the location of the crops (\textit{geometric pooling} \cite{ziegler2022self,xiao2021region}) and/or an attention mechanism (\textit{attentive pooling} \cite{ziegler2022self,xie2021propagate,o2020unsupervised,wen2022self,wang2022exploring}) can be used to infer the region of overlap in each view and only apply the consistency objective to that region (Fig.~\ref{fig:joint_vs_intersect}\textcolor{red}{.c}). A consequence of these pooling mechanisms is that only a sub-region of each view is exploited, which mislays a significant amount of the image and further questions the usage of \textit{cropping}. There are two strategies to tackle the issue of locating and linking the objects from the two views: the first is a feature-level approach that extends the global consistency criterion to the spatial features after inferring pairs of positives through similarity bootstrapping or positional cues \cite{liu2020self,xie2021propagate,bardes2022vicregl,wang2021dense,li2021efficient,ziegler2022self}. It is unclear how much semantics a single spatial feature embeds, and this strategy can become computationally intensive. These issues motivate the emergence of the second line of work which operates at the object-level \cite{wen2022self,henaff2021efficient,henaff2022object,xie2021unsupervised,van2021unsupervised,wang2022freesolo,wei2021aligning}. In that second scenario, the main difficulty lies in generating the object segmentation masks and matching objects from one view to the other. The straightforward approach is to leverage unsupervised heuristics \cite{henaff2021efficient} or pre-trained models \cite{xie2021unsupervised} to generate pseudo labels prior to the training phase (Fig.~\ref{fig:joint_vs_intersect}\textcolor{red}{.a}), which is not an entirely data-driven approach and cannot be trivially extended to any modalities. Alternatively, \cite{henaff2022object} proposed to use K-Means and an additional global image (encompassing the two main views) to generate online pseudo labels, but this approach is computationally intensive. \par
To address these limitations, we propose $\mname$, whose underpinning mechanism is an efficient \textbf{Cr}oss-view \textbf{O}nline \textbf{C}lustering that conjointly generates segmentation masks for the union of both views (Fig.~\ref{fig:joint_vs_intersect}\textcolor{red}{.b}).

Our main contributions are: 1) we propose a novel object-level self-supervised learning framework that leverages an online clustering algorithm yielding segmentation masks for the union of two image views. 2) The introduced method is inherently compatible with scene-centric datasets and does not require a pre-trained model. 3) We empirically and thoroughly demonstrate that our approach rivals or out-competes existing SOTA self-supervised methods even when pre-trained in an unfavorable setting (smaller and more complex dataset). 
\section{Related work}
\label{sec:related_work}
\noindent
{\bf Global features.}
\label{par:global_features}
The collateral effect of \cite{chen2020simple}, is that it effectively uniformized the choice of the proxy task for SSL to the extent that \textit{cross-view consistency} is almost exclusively used. The remaining degree of freedom lies in the technique used to avoid the collapse to trivial solutions. The use of negative samples \cite{chen2020simple,hjelm2018learning} effectively and intuitively treats this degeneracy at the cost of using large batches, which can be mitigated by a momentum encoder \cite{he2020momentum}. At the other end of the spectrum, clustering-based approaches \cite{asano2019self,caron2018deep,caron2020unsupervised,caron2021emerging} have shown that enforcing equipartition of the samples over a set of clusters was sufficient to palliate the collapsing issue.

\noindent
{\bf Local features.}
\label{par:local_features}
Local methods aim at completing the image-level objective by encouraging cross-view consistency at a localized level such that the resulting features are well aligned with dense downstream tasks. Broadly speaking, these methods can be categorized by the granularity at which the similarity is enforced. The first category encompasses approaches \cite{wang2021dense,liu2020self,o2020unsupervised,lebailly2022global}, where similarity is encouraged directly at the feature level, i.e., from one feature to the other. The difficulty lies in obtaining valid pairs or groups of features. To that end, various methods \cite{wang2021dense,liu2020self} rely solely on the similarity of the features, whereas the matching criterion of \cite{xie2021propagate,o2020unsupervised} is driven by their distances/positions. 
\cite{lebailly2022global} studies both approaches and \cite{bardes2022vicregl} incorporates both in a single objective.\par

The second category of methods \cite{cho2021picie,wen2022self,henaff2021efficient,henaff2022object,xie2021unsupervised} enforce consistency at a higher level, which first requires finding semantically coherent groups of features. For that purpose, \cite{xie2021unsupervised}, resort to using a pre-trained model and an offline ``correspondences discovery'' stage to find pairs of the region of interest. Along the same line, \cite{henaff2021efficient} proposes to use various heuristics prior to the training phase to generate pseudo-segmentation labels. An online version of this latest algorithm has been introduced, but it requires forwarding an additional global view. 
\par

Alternatively, dense fine-tuning approaches \cite{hamilton2022unsupervised,ziegler2022self,yun2022patch,wang2022cp2} have been proposed. These methods aim to endow models pre-trained under an image-level objective \cite{caron2021emerging} with local consistency properties, but cannot be trained from scratch.
\par
Finally, MAE \cite{he2021masked} relies on a masked autoencoder pipeline and a reconstruction objective to learn dense representations. As MAE does not rely on a cross-view consistency objective, this approach is well-suited for scene-centric datasets and of particular interest to us.


\section{Method}
\label{sec:method}
\begin{figure*}[ht]
    \centering
    \includegraphics[width=\textwidth]{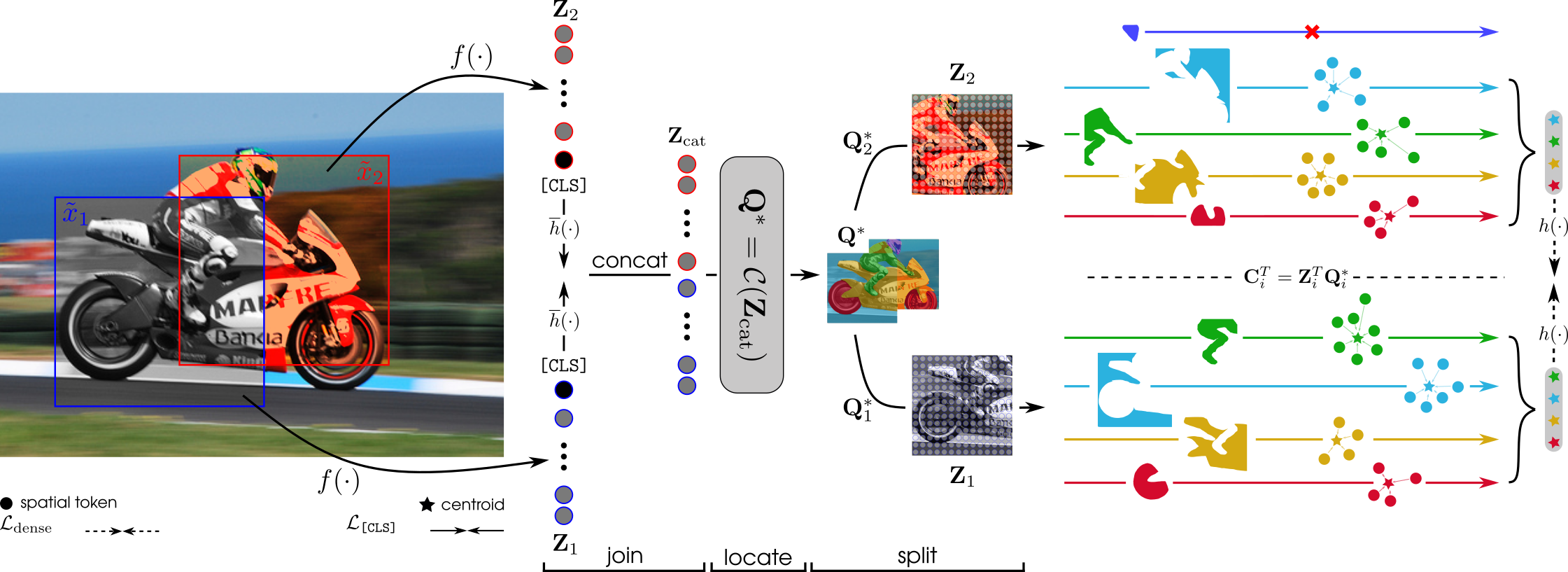}
    \caption{
    \textbf{Overview of $\mname$.} The augmented views, $\aug_1$ and $\aug_2$, are processed independently by a ViT encoder $\backbone$. The \textit{joint} representation, $\dense_{\text{cat}}$, of the two image views, is obtained by concatenation along the token axis and serves as input to the clustering algorithm, $\mathcal{C}$, to \textit{locate} the objects. The joint clustering assignments, $\Q^{*}$, are \textit{split} view-wise and used to compute the corresponding centroids. A self-distillation loss enforces consistency between pairs of related centroids via a projection head $h$.}
    \label{fig:pipeline}
\end{figure*}
\subsection{Overview}
\label{ssec:method_overview}
This paper tackles the problem of learning dense visual representations from unlabeled scene-centric data. Recent efforts using a self-supervised multi-view consistency paradigm to address this problem rely on a two steps procedure: \textit{i)} \textit{locate} the objects in each image view and \textit{ii)} \textit{link} the related objects from one image view to the other. We now discuss how $\mname$ elegantly palliates the limitations evoked in sections~\ref{sec:introduction} and~\ref{sec:related_work}.

\par
We observe that most of the difficulties arise because the \textit{locate-link} strategy treats the two image views independently. In contrast, both views stem from the same image, and their representations lie in the same space. The former observation offers the possibility to benefit from the coordinates of the cropped image regions as a cue for the \textit{locate} step, while the latter indicates that some operations could be performed conjointly. Consequently, we propose to depart from the typical strategy and introduce a novel paradigm dubbed \textit{join-locate-split}, described below:

\par \noindent
\textbf{Join.}
The two augmented image views, $\aug_1$ and $\aug_2$, are processed by a ViT \cite{dosovitskiy2020image} encoder $\backbone$ yielding the dense visual representations $\dense_{\{1,2 \}} \in \mathbb{R}^{N \times d}$, where $N$ and $d$ denote the number of spatial tokens and feature dimension, respectively. The dense visual representations are then concatenated along the token axis to obtain the joint representation, $\dense_{\text{cat}} \in \mathbb{R}^{2N \times d}$.
 \par\noindent
\textbf{Locate.}
The objective is to find semantically coherent clusters of tokens in the joint representation space. As the quality of the input representation improves, we expect the found clusters to represent the different objects or object parts illustrated in the image. The joint representation is fed to the clustering algorithm $\mathcal{C}$, which outputs the joint clustering assignments, $\Q^{*} \in \mathbb{R}^{2N \times K}$. The soft assignments matrix $\Q^{*}$ models the probability of each of the $2N$ tokens to belong to one of the $K$ clusters found in the joint space.  
 \par \noindent
\textbf{Split.}
By splitting $\Q^{*}$ in two along the first dimension, the assignment matrix of each view, namely $\Q^{*}_{\{1,2 \}} \in \mathbb{R}^{N \times K}$ are obtained. One can observe that the \textit{link} operation is provided for free and that it is trivial to discard any cluster that does not span across the two views.
\par
Given the view-wise assignments $\Q^{*}_{\{1,2 \}}$, and the corresponding dense representations $\dense_{\{1,2 \}}$, $K$ object/cluster-level representations can be obtained for each view:
\begin{align}
    \label{eq:compute_cetntroids}
    \begin{split}
    \centroids^{\top}_{1} = \dense^{\top}_{1} \Q^{*}_{1}
    \end{split}
\end{align}
%
$\centroids$ denotes the centroids. Analogously to the image-level consistency objective, one can enforce similarity constraints between pairs of centroids.

\subsection{Dense self-distillation}
\label{ssec:dense_self_distillation}
This section details the integration of the \textit{join-locate-split} strategy (Sec.~\ref{ssec:method_overview}) in a self-distillation scheme\footnote{Our implementations build upon DINO \cite{caron2021emerging}, but it's not limited to it.}.
Our self-distillation approach relies on a teacher-student pair of Siamese networks, $\model_t$ and $\model_s$, each composed of an encoder $\backbone_{\{t,s \}}$ and a projection head $\head_{\{t,s\}}$. Given the input image $\im \in \mathbb{R}^{C \times H \times W}$, two augmented views $\aug_{1}$ and $\aug_{2}$ are obtained using random augmentations. Both augmented views are independently passed through the teacher and student encoders, yielding the spatial representations $\dense_{t,\{1,2\}}$ and $\dense_{s,\{1,2\}}$, respectively. The teacher model's representations are concatenated (\textit{join}) and fed to the clustering algorithm (Sec.~\ref{ssec:where_objects?}) to obtain the assignment matrix $\Q^{*}$ (\textit{locate}), which is assumed to be already filtered of any column corresponding to an object/cluster represented in only one of the two views (cf. Sec.~\ref{sssec:pruning}). The assignment matrix is \textit{split} view-wise to get $\Q^{*}_{\{1,2 \}}$ and to compute the teacher and student centroids of each view:
\begin{align}
\begin{split}
        \centroids^{\top}_{\{t,s\},\{1,2 \}} &= \dense^{\top}_{\{t,s \},\{1,2\}} \Q^{*}_{\{1,2\}} \\
\end{split}
\end{align}
The final step is to feed the teacher and student centroids, $\centroids_{t}$ and $\centroids_{s}$, to the corresponding projection heads, $\head_{t}$ and $\head_{s}$, which output probability distributions over $L$ dimensions denoted by $\PP_{t}$ and $\PP_{s}$, respectively. The probabilities of the teacher and student models are obtained by normalizing their projection heads' outputs with a \texttt{softmax} scaled by  temperatures $\tau_{t}$ and $\tau_{s}$:
\begin{align}
    \label{eq:projection}
    \begin{split}
    \PP_{t, \{1,2 \}} &= \underset{L}{\texttt{softmax}}\left(\head_{t}(\centroids_{t,\{1,2\}}) / \tau_{t} \right)  \\
    \PP_{s, \{1,2 \}} &= \underset{L}{\texttt{softmax}}\left(\head_{s}(\centroids_{s,\{1,2\}}) / \tau_{s} \right)  \\
    \end{split}
\end{align}
The dense self-distillation objective $\mathcal{L}_{\text{dense}}$ enforces cross-view consistency of the teacher and student model projections using the cross-entropy loss:
\begin{equation}
    \label{eq:dense_loss}
\mathcal{L}_{\text{dense}} = \frac{1}{2}\left( H(\PP_{t, 1}, \PP_{s,2}) +  H(\PP_{t, 2}, \PP_{s,1}) \right)
\end{equation}
where $H(\mathbf{A}, \mathbf{B}) = -\frac{1}{K} \sum_{k=1}^{K} \sum_{l=1}^{L} \mathbf{A}_{kl} \log (\mathbf{B}_{kl})$ computed by averaging over all clusters.\par

For the dense self-distillation loss to be meaningful, the clustering assignments of spatial tokens corresponding to similar objects must be semantically coherent, which requires good-quality representations. To address this issue, we additionally apply a global representation loss by feeding the image-level representations to a dedicated projection head, $\overbar{\head}$, to obtain the $\overbar{L}$-dimensional distributions:


\begin{align}
    \label{eq:image_proj}
    \begin{split}
    \p_{t, \{1,2 \}} &= \underset{\overbar{L}}{\texttt{softmax}} \left(\overbar{\head}_{t}(\glob_{t,\{1,2\}} / \overbar{\tau}_{t})\right) \\
    \p_{s, \{1,2 \}} &= \underset{\overbar{L}}{\texttt{softmax}} \left(\overbar{\head}_{s}(\glob_{s,\{1,2\}} / \overbar{\tau}_{s})\right)
    \end{split}
\end{align}
The sharpness of the output distribution for teacher and student models is controlled by the temperature parameters $\overbar{\tau}_{t}$ and $\overbar{\tau}_{s}$, respectively, and $\glob_{\{t, s\}}$ denotes the image-level representations of the teacher and student models. Hence the global representation loss $\mathcal{L}_{\text{glob}}$ is computed as follows:
\begin{equation}
    \label{eq:glob_loss}
    \mathcal{L}_{\text{glob}} =  \frac{1}{2}\left( H(\p_{t, 1}, \p_{s,2}) +  H(\p_{t, 2}, \p_{s,1}) \right)
\end{equation}
where $H(\mathbf{a}, \mathbf{b}) = -\sum_{l=1}^{\overbar{L}} \mathbf{a}_{l} \log (\mathbf{b}_{l})$. Therefore, the overall loss function used for the training of $\mname$ is:
\begin{align}
    \label{eq:total_loss}
    \mathcal{L} = \alpha \mathcal{L}_{\text{dense}} + \mathcal{L}_{\text{glob}}
\end{align}
where $\alpha$ denotes a hyperparameter to balance the loss terms. We set $\alpha=1.0$ for all experiments without the need for hyperparameter tuning.

\subsection{Where are the objects in the image?} 
\label{ssec:where_objects?}
\begin{figure*}
    \centering
    \includegraphics[width=\textwidth]{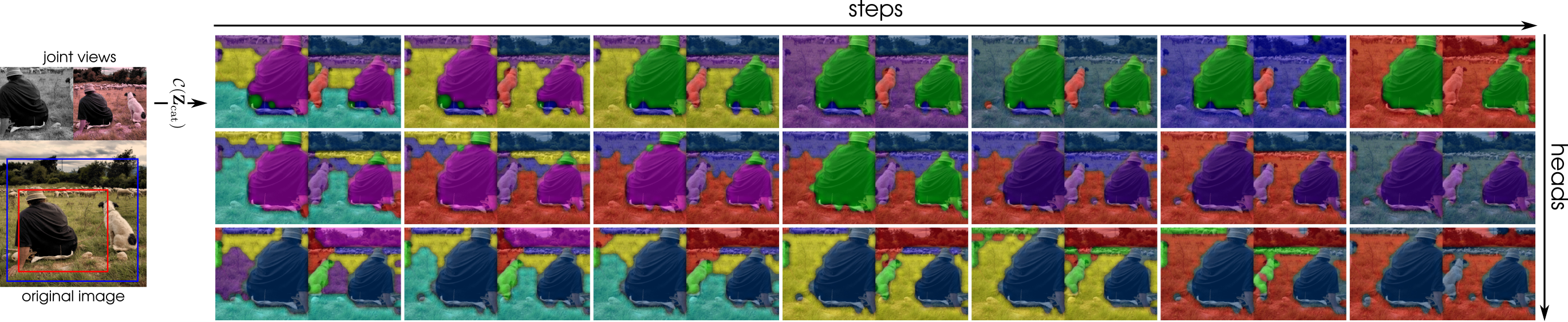}
    \caption{
    \textbf{Representation of the iterative clustering algorithm in the joint space.} The algorithm is initialized with a fixed number of centroids that are iteratively merged until only two remain. The ideal number of centroids is determined \textit{a posteriori}. The procedure's last seven steps (columns) are represented for three different heads of a ViT-S/16 pre-trained with \mname. 
    }
    \label{fig:segmentation_masks}
\end{figure*}
So far, we assumed that there existed an algorithm able to assign a set of input data points to an undetermined number of clusters. This section covers the details of this algorithm.\par
The online optimization objective for computing the clusters and corresponding assignments relies on an optimal transport formulation and the Sinkhorn-Knopp algorithm \cite{cuturi2013sinkhorn}. This choice is motivated by \textit{i)} its efficiency, \textit{ii)} the ease of incorporating  external knowledge (Sec.~\ref{sssec:position_cues}), and \textit{iii)} it returns a measure of the clustering quality, which can be used to infer the optimal number of clusters $K$ for a given image. The last point is of utmost importance as it allows us to devise a \textit{ad-hoc} selection criterion for $K$. Indeed, the iterative procedure progressively merges the centroids until only two remain, i.e., background/foreground (see Fig.~\ref{fig:segmentation_masks}). The number of centroids $K$ is selected \textit{a posteriori} and independently for each image in the batch.
\par
More formally, let's consider a ViT encoder $\backbone$ fed with a positive pair of augmented views, $\aug_{1}$ and $\aug_{2}$, and yielding the corresponding representations, $\dense_{1}$ and $\dense_{2}$. The clustering is performed on the joint representation, $\dense_{\text{cat}} \in \mathbb{R}^{2N \times d}$ obtained from the concatenation of $\dense_{1}$ and $\dense_{2}$ along the token axis. The procedure starts by sampling $K_{\text{start}}$ of the $2N$ tokens, which serve as initialization for the centroids, $\centroids \in \mathbb{R}^{K_{\text{start}} \times d}$:
\begin{equation}
    \label{eq:cent_init}
    \centroids = \Y^{\top} \dense_{\text{cat}}
\end{equation}
where $\Y \in \{0,1\}^{2N \times K_{\text{start}}}$ is a matrix of column one-hot vectors indicating the position of the $K_{\text{start}}$ tokens used to initialize the centroids.
The sampling is based on the \textit{attention map} of the \texttt{[CLS]} token, which highlights the patches proportionally to their contribution to the image-level representation. The cost of assigning a token to a given centroid should reflect their similarity, hence:
 \begin{equation}
     \label{eq:cost}
     \costsem = - \dense_{\text{cat}} \centroids^{\top}
 \end{equation}
where $\costsem \in \mathbb{R}^{2N \times K}$ denotes the cost matrix of the assignments. A handy property of the selected clustering algorithm is that it offers the possibility to scale the importance of the tokens and centroids based on external knowledge injected using a token distribution $\tmarg$ and a centroid distributions $\cmarg$. Here, the \textit{attention map} of the \texttt{[CLS]} token is used as the token distribution due to its ability to highlight the sensible semantic regions of the image \cite{caron2021emerging}. Along the same line, the centroids distribution is defined as:
 \begin{equation}
     \label{eq:cmarg}
     \cmarg =  \texttt{softmax}(\Y^{\top} \tmarg)
 \end{equation}
%
Given the cost matrix, $\costsem$, and the two marginals, $\tmarg$ and $\cmarg$, the Sinkhorn-Knopp clustering produces the assignment matrix $\Q^{*}$:

\begin{equation}
    \label{eq:assignments}
    \Q ^{*} = \underset{\Q \in \mathcal{U}(\tmarg, \cmarg)}{\arg\min} <\Q, \costsem> - \frac{1}{\lambda} H(\Q)
\end{equation}
where $<\cdot, \cdot >$ denotes the entry-wise product followed by a sum reduction. The second term is a regularization of the entropy of the assignments, i.e., it controls the sharpness of the clustering. $\mathcal{U}(\tmarg, \cmarg)$ is the transportation polytope, i.e., the set of valid assignments defined as:
\begin{equation}
    \label{eq:transport_polytope}
    \mathcal{U}(\tmarg, \cmarg) = \{ \Q \in \mathbb{R}^{2N \times K}_{+} \:|\: \Q \mathbf{1}_{K} = \tmarg, \Q^{\top} \mathbf{1}_{2N} = \cmarg \}
\end{equation}
Additionally, the transportation cost $d_{\text{c}}$ measures the cost of assigning the tokens to the different centroids and can therefore be interpreted as the quality of the clustering, i.e., the ability to find a representative centroid for each token.
\begin{equation}
\label{eq:transportationcost}
    d_{\text{c}} = \: <\Q^{*}, \costsem>
\end{equation}
%
The centroids are updated after each step ($\centroids^{\top} = \dense^{\top} \Q^{*}$),
%
%
and the two centroids, $(i^{*}, j^{*})$, having the highest cosine similarity, are merged:
\begin{equation}
\label{eq:merge}
    \centroids, \Y \leftarrow \texttt{merge}(\centroids, \Y, i^{*}, j^{*})
\end{equation}
where \texttt{merge} denotes the merging operator; the merging procedure averages the selected columns of $\Y$ and the corresponding rows of $\centroids$; in both cases, obsolete columns/rows are simply removed from the matrices. Before reiterating through the clustering algorithm, the matrix cost, $\costsem$, and centroid distribution, $\cmarg$, are updated using Eq.~\ref{eq:cost} and Eq.~\ref{eq:cmarg}, respectively. The whole procedure is repeated until only two centroids remain. By comparing the transportation cost $d_{\text{c}}$ incurred at each step (from $K_{\text{start}}$ to 2), one can select \textit{a posteriori} the optimal number of centroids and the corresponding assignment $\Q^{*}$ for each image independently based on the $\Q^{*}$ that minimizes $d_{\text{c}}$. The procedure's final step consists of the row-wise normalization of the assignments and the pruning of clusters (cf. Sec.~\ref{sssec:pruning}).
\subsubsection{Cluster pruning}
\label{sssec:pruning}

An important property of $\mname$ is that it allows to easily discard clusters corresponding to content that is not shared across the two views (e.g., purple cluster corresponding to the helmet in Fig.~\ref{fig:pipeline}). To that end, we first compute the hard version of the assignments (each token is assigned to precisely one centroid):

\begin{equation}
\label{eq:binary_mask}
    \mask_{n,k} = \mathds{1}_{ k = \underset{j}{\text{argmax}}\left\{ \Q^{*}_{nj}\right\}}
\end{equation}
The hard assignments are split view-wise to obtain $\mask_{1}$ and $\mask_{2}$, and we introduce the sets $\mathcal{S}_{1}$ and $\mathcal{S}_{2}$, which store indices of the zero columns of $\mask_{1}$, and  $\mask_{2}$, respectively. Therefore, any column of $\Q^{*}_{\{1,2\}}$ and $\mask_{\{1,2 \}}$, whose index is in $\mathcal{S} = \mathcal{S}_{1} \cup \mathcal{S}_{2}$, is filtered out:
 %
\begin{equation}
\label{eq:filter}
    \Q_{\{1,2\}}^{*}, \mask_{\{1,2 \}} \leftarrow \texttt{filter}(\Q_{\{1, 2\}}^{*}, \mask_{\{1,2 \}}, \mathcal{S})
\end{equation}
where \texttt{filter} denotes the filtering operator, which drops the indexed columns of the input matrices. 
\subsubsection{Positional cues}
\label{sssec:position_cues}
In Sec.~\ref{ssec:dense_self_distillation}, we mention the need for an image-level self-distillation loss to break the interdependence between the features' quality and the correctness of the enforced dense loss.  Along the same line, positional cues can be leveraged to guide the clustering operation, such that spatially coherent clusters can be obtained even when the features do not fully capture the semantics of the underlying data. Indeed, it appears natural to bias the clustering in favor of matching together tokens resulting from the same region \textbf{in the original image}. To that end, a positional constraint is added to the matrix transportation cost $\costsem$, which is modified to incorporate this desired property.
\begin{figure}[h]
    \centering
    \includegraphics[width=0.5\columnwidth]{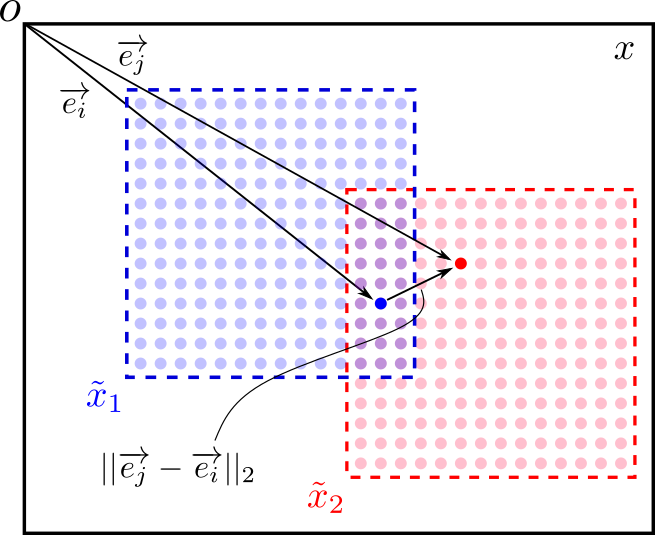}
    \caption{
    The positional cues use the top-left corner of the \textbf{original image} as a reference point, such that the position coordinates of each view lie in the same space and can be used to guide the clustering algorithm.
    }
    \label{fig:positional_cues}
\end{figure}
\par

We start by observing that the augmented views, $\aug_{\{1,2\}}$, result from the composition and use of a set of geometric and photometric transformations on the original image $\im$. We propose to extract the coordinates of the patches in each view with respect to the original image referential (cf. Fig.~\ref{fig:positional_cues}). More precisely, we generate the tensors, $\pos_{\{1,2\}} \in \mathbb{R}^{N\times 2}$, which store the 2D coordinates of each patch in the two views. 
The coordinates are first concatenated along the patch/token axis to obtain $\pos_{\text{cat}} \in \mathbb{R}^{2N \times 2}$, and the positions of the centroids, $\pos_{\text{cen}} \in \mathbb{R}^{K_{\text{start}} \times 2}$, are computed as in Eq. \ref{eq:cent_init} ($\pos_{\text{cen}} = \Y^{\top} \pos_{\text{cat}}$). The entries of the positional transportation cost $\costpos \in \mathbb{R}^{2N \times K_{\text{start}}}$ are computed as follows:
\begin{equation}
    \costpos_{ij} = \frac{1}{S}|| \mathbf{e}^{\text{(cat)}}_{i} -  \mathbf{e}^{\text{(cen)}}_{j}||_{2}
\end{equation}
where $S$ is a normalization constant that ensures that the entries of the positional transportation cost are in $[0, 1]$. After incorporation of the positional bias, the total matrix transportation cost is defined as follows:
\begin{equation}
\costtot = \costsem + \lambda_{\text{pos}} \costpos
\end{equation}
The scalar weight  $\lambda_{\text{pos}}$ regulates the importance of the positional cues. As detailed in Sec.~\ref{ssec:where_objects?}, the clustering algorithm relies on the iterative merging of the centroids; hence their respective position must also be merged reciprocally, i.e., by averaging (cf. Eq.~\ref{eq:merge}).

\subsubsection{Multiple clustering assignments using MSA}
\label{sssec:multiple_heads}
In this section, we detail a mechanism to obtain multiple complementary clustering assignments $\Q^{*}$ per image. This mechanism relies on the multi-head self-attention (MSA) module inherent to the transformer architecture. \par
Arguably, the main ingredient behind the transformer architecture's success is the self-attention module. Indeed, \textit{i)} it allows capturing of long-range inter-dependencies between the patches that constitute the image, and \textit{ii)} it endows the local representations with global or contextual information. Formally, the multi-head attention operation of the $l^{th}$ transformer block is expressed as:
\begin{align}
\label{eq:multi_head}
\begin{split}
\texttt{multi}&\texttt{-head} \left(\dense^{(l-1)} \right) \\
 &= \texttt{concat}\left(\text{head}_{1}, ...,\text{head}_{n_{h}}\right)\mathbf{W}^{o}
\end{split}
\end{align}
where $\mathbf{W}^{o} \in \mathbb{R}^{d \times d}$ is a learnable projection weight, and $\text{head}_{i}$, for $i=1,\cdots ,n_{h}$, denotes a single attention head:
\begin{align}
\label{eq:single_head} 
\text{head}_{i} &= \texttt{attention}\left(\dense^{(l-1)}, \mathbf{W}^{\{q,k,v\}}_{i} \right) \\
 &= \texttt{softmax}\left( \frac{\dense^{(l-1)} \mathbf{W}_{i}^{q} \left(\dense^{(l-1)} \mathbf{W}_{i}^{k}\right)^{\top}}{\sqrt{D}} \right) \dense^{(l-1)} \mathbf{W}_{i}^{v} \nonumber
\end{align}
where $\mathbf{W}_{i}^{\{q,k,v \}} \in \mathbb{R}^{d \times d/n_{h}}$ denotes head-specific learnable projection weights\footnote{The layer index, which starts from 0, is omitted.}. Following the same reasoning that motivates the use of multiple heads, i.e., the inter-patches relationship is not unique, we use as many clustering assignments as there are heads in the MSA module. In practice, it turns out to be as simple as independently feeding each attention head's dense representation to the clustering algorithm:
\begin{equation}
    \Q^{i} = \mathcal{C}\left(\dense^{(B-2)} \mathbf{W}_{i}^{k} \right)
\end{equation}
where $B$ is the number of transformer blocks in the model. Note that only one of the \texttt{keys/queries/values} representation is used (here exemplified with the \texttt{keys}). Consequently, the final assignment matrix $\Q^{*}$ results from the concatenation of the head-wise assignments $\Q^{i}$
%
along the centroid dimension. Up to pruning (Sec.~\ref{sssec:pruning}), the effective number of centroids is $n_{h}$ times higher. Even though the clusters overlap, we do not enforce contradictory objectives as \textit{i)} the consistency is enforced pair-wise (from one centroid in the first view to the corresponding one in the second view) and \textit{ii)} in the framework of self-distillation there are no negative pairs.
\section{Experiments}
\label{sec:experiments}

\subsection{Implementations details}
\label{ssec:implementation_details}
\noindent
{\bf Pre-training datasets.}
\label{par:pre_training}
Our models are pre-trained on two uncurated and scene-centric datasets, namely COCO (\texttt{train2017}, $\sim$118k images) and COCO+ (\texttt{unlabeled2017} + \texttt{train2017}, $\sim$241k images). We further explore the possibility of using $\mname$ in an object-centric scenario and therefore adopt ImageNet \cite{5206848} as a pre-training dataset ($\sim10\times$ more images and $\sim4\times$ fewer objects/image).

\noindent
{\bf Network architecture.}
\label{par:network}
We use a ViT-small (ViT-S/16) as the backbone $f$. This choice is in line with its adoption in concurrent methods and for its comparability \cite{caron2021emerging,ziegler2022self,yun2022patch} with the $\resnet$50, which is the backbone of the remaining baselines. The architecture of the projection heads is identical to that of \cite{caron2021emerging}. Notably, the image-level and centroids-level heads, $\overbar{\head}$ and $\head$, share their weights except for the last layer, which has output dimensions, $\overbar{L}=65,536$ and $L=8,192$, respectively.

\noindent
{\bf Optimization.}
\label{par:optimization}
 $\mname$ is trained for 300 epochs on COCO and COCO+ under an identical optimization scheme. A batch size of 256, distributed over 2 Tesla V100 GPUs is used. The pre-training on ImageNet uses a batch size of 1024, distributed over 4 AMD MI250X GPUs. The remaining optimization setting is identical to that of $\dino$ \cite{caron2021emerging}. 

\noindent
{\bf Hyperparameters.}
The same weight is given to the dense and global loss, i.e., $\alpha=1$. We use $\lambda=20$ for the regularization term of the transportation objective. The dense and global projection heads use the same temperature parameters, namely $\overbar{\tau}_{s} = \tau_{s} = 0.1$ and $\overbar{\tau}_{t} = \tau_{t} = 0.07$ (see Eqs.~\ref{eq:projection} \&~\ref{eq:image_proj}). Generally, any hyper-parameter common to $\dino$ uses its recommended value. The results of section~\ref{ssec:segmentation_results} which use COCO or COCO+ as pre-training datasets are obtained with $\lambda_{\text{pos}}=4$, $K_{\text{start}}=12$ and the $\texttt{values}$ tokens as parameters of the clustering algorithm. For ImageNet, we only report results with $\lambda_{\text{pos}}=3$, $K_{\text{start}}=12$ and the $\texttt{keys}$ tokens. These values correspond to the default setting of the grid search performed on COCO (see~\cref{ssec:ablation}).

\subsection{Evaluation protocols}
\label{ssec:evaluation_protocols}
We opt for dense evaluation downstream tasks, which require as little manual intervention as possible, such that the reported results truly reflect the quality of the features. Details of the implementations and datasets are available in~\Cref{sec:app_evaluation_protocols}.

\noindent
{\bf Transfer learning via linear segmentation.}
\label{par:linear_segmentation}
The linear separability of the learned spatial features is evaluated by training a linear layer on top of the frozen features of the pre-trained encoder. The linear layer implements a mapping from the embedding space to the label space and is trained to minimize the cross-entropy loss. We report the mean Intersection over Union (mIoU) of the resulting segmentation maps on four different datasets, namely, PVOC12 \cite{pascal-voc-2012}, COCO-Things, COCO-Stuff \cite{lin2014microsoft} and ADE20K \cite{zhou2017scene}.

\noindent
{\bf Transfer learning via unsupervised segmentation.}
\label{par:kmeans_segmentation}
We evaluate the ability of the methods to produce spatial features that can be grouped into coherent clusters. We perform  K-Means clustering on the spatial features of every image in a given dataset with as many centroids as there are classes in the dataset. Subsequently, a label is assigned to each cluster via Hungarian matching \cite{kuhn1955hungarian}. We report the mean Intersection over Union (mIoU) of the resulting segmentation maps on three different datasets, namely PVOC12 \cite{pascal-voc-2012}, COCO-Things, and COCO-Stuff \cite{lin2014microsoft}.
\noindent
{\bf Semi-supervised video object segmentation.}
\label{par:video_segmentation}
We assess our method's generalizability for semi-supervised video object segmentation on the DAVIS'17 benchmark. The purpose of this experiment is to evaluate the spatiotemporal consistency of the learned features. First, the features of each frame in a given video are independently obtained; secondly, a nearest-neighbor approach is used to propagate (from one frame to the next) the ground-truth labels of the first frame  (see results in~\cref{sec:app_video_segmentation}).

\subsection{Segmentation results}
\label{ssec:segmentation_results}
In~\Cref{table:linear_segmentation,table:linear_segmentation_ade20k}, we report mIoU results on the linear segmentation task. When pre-trained on COCO, $\mname$ exceeds concurrent methods using COCO(+) as pre-training datasets, even though $\orl$ and $\byol$ use a longer training protocol (800 epochs). With a pre-training on COCO+, $\mname$ outperforms all other methods, except $\cpsq$ \cite{wang2022cp2}, on every evaluation dataset, despite their usage of ImageNet and their finetuning on one of the target datasets (PVOC12). Noteworthy that $\cpsq$ is initialized with a pre-trained model and cannot be trained from scratch. Pre-training on a larger and object-centric dataset appears to be highly beneficial in that setting.
 \par
\begin{table}[!ht]
\centering
\caption{
\textbf{Transfer results of linear segmentation task.} A linear layer is trained on top of the frozen spatial features. The mIoU scores are reported on the PVOC12 \cite{pascal-voc-2012}, COCO-Things (CC-Th.), and COCO-Stuff (CC-St.) \cite{lin2014microsoft}. The pre-training dataset is either of ImageNet (IN) \cite{5206848}, COCO (CC), or COCO+ (CC+).
}
\footnotesize
\resizebox{1\columnwidth}{!}{
\begin{tabular}{l c c c c c c c c c}
\toprule
Method & Model / Dataset  & PVOC12 & CC-Th. & CC-St. & Avg.\\
\midrule
\textit{Global features} \\
$\byol$ \cite{grill2020bootstrap} & $\resnet$50 / CC+     & 38.7 & 50.4 & 39.8 & 43.0 \\
$\dino$ \cite{caron2021emerging} & $\vit$-S/16 / CC       & 47.2 & 47.1 & 46.2 & 46.8 \\
\midrule
\textit{Local features} \\
$\orl$ \cite{xie2021unsupervised} & $\resnet$50 / CC+     & 45.2 & 55.6 & 45.6 & 48.8 \\
$\densecl$ \cite{wang2021dense} & $\resnet$50 / IN & 57.9 & 60.4 & 47.5 & 55.3 \\
$\soco$ \cite{wei2021aligning} & $\resnet$50 / IN  & 54.0 & 56.8 & 44.2 & 51.7 \\
$\resim$ \cite{xiao2021region} & $\resnet$50 / IN  & 55.1 & 57.7 & 46.5 & 53.1 \\
$\pixpro$ \cite{xie2021propagate} & $\resnet$50 / IN & 57.1 & 54.7 & 45.9 & 52.6\\
$\vicregl$ \cite{bardes2022vicregl} & $\resnet$50 / IN & 58.9 & 58.7 & 48.2 & 55.3 \\
$\mae$ \cite{he2021masked} & $\vit$-S/16 / CC    & 31.7 & 35.1 & 39.6 & 35.5 \\
$\cpsq$ \cite{wang2022cp2} & $\vit$-S/16 / IN+PVOC12 & \underline{63.1} & 59.4 & 46.5 & 56.3 \\
\midrule
\textit{Ours} \\
\rowcolor{cyan!20} $\mname$ & $\vit$-S/16 / CC & 54.5 & 55.6 & 49.7 & 53.3 \\
\rowcolor{cyan!20} $\mname$ & $\vit$-S/16 / CC+ & 60.6 & \underline{62.7} & \underline{51.7} & \underline{58.3} \\
\rowcolor{cyan!20} $\mname$ & $\vit$-S/16 / IN & \textbf{70.6} & \textbf{66.1} & \textbf{52.6} & \textbf{63.1} \\
\bottomrule
\end{tabular}
\label{table:linear_segmentation}
}
\end{table}%



In Table~\ref{table:kmeans_segmentation}, the results for the unsupervised segmentation task are reported. As for linear segmentation experiments, $\mname$ is already competitive with only a pre-training on the COCO dataset and surpasses all competing methods except $\densecl$ \cite{wang2021dense}. 
\begin{table}[!ht]
\centering
\caption{
\textbf{
Transfer results of linear segmentation task.} A linear layer is trained on top of the frozen spatial features. The mIoU scores are reported for ADE20k \cite{zhou2017scene}. The pre-training dataset is either ImageNet \cite{5206848}, COCO, or COCO+.
}
\footnotesize
\resizebox{1\columnwidth}{!}{
\begin{tabular}{l c c c c c c c c c}
\toprule
Method & Model & Dataset  & Epochs & mIoU \\
\midrule
\textit{Global features} \\
$\dino$ \cite{caron2021emerging} & $\vit$-S/16 & COCO & 300 & 18.5 \\
$\dino$ \cite{caron2021emerging} & $\vit$-S/16 & ImageNet & 800 & 26.8  \\
\midrule
\textit{Local features} \\
$\densecl$ \cite{wang2021dense} & $\resnet$50 & ImageNet & 200 & 24.3  \\
$\vicregl$ \cite{bardes2022vicregl} & $\resnet$50 & ImageNet & 300 & 23.7 \\
$\cpsq$ \cite{wang2022cp2} & $\vit$-S/16 & ImageNet+PVOC12 & 320 & 25.4 \\
\midrule
\textit{Ours} \\
\rowcolor{cyan!20} $\mname$ & $\vit$-S/16 & COCO & 300 & 23.2   \\
\rowcolor{cyan!20} $\mname$ & $\vit$-S/16 & COCO+ & 300 & \underline{27.0}  \\
\rowcolor{cyan!20} $\mname$ & $\vit$-S/16 & ImageNet & 300 & \textbf{28.4}  \\
\bottomrule
\end{tabular}
\label{table:linear_segmentation_ade20k}
}
\end{table}%

The largest improvements are observed on the COCO-Stuff dataset; this is unsurprising as this dataset contains semantic labels such as $\texttt{water}$, $\texttt{ground}$, or $\texttt{sky}$, which correspond to regions that are typically overlooked by other methods, but on which $\mname$ puts a significant emphasis. The model pre-trained on ImageNet appears to perform poorly on that task, which is surprising considering the excellent results depicted in~\Cref{table:linear_segmentation} on the exact same datasets. This might hint that using evaluations that are not adjustable to each baseline is sub-optimal. Overall we observe that producing features that can be clustered class-wise without labels remains an open challenge.
\begin{table}[!ht]
\centering
\caption{
\textbf{Transfer results of unsupervised segmentation task.} The frozen spatial features of each image in a given dataset are clustered into as many clusters as there are classes in the dataset. The Hungarian matching algorithm is used to label the clusters. The mIoU scores are reported on PVOC12 \cite{pascal-voc-2012}, COCO-Things (CC-Th.) and COCO-Stuff (CC-St.) \cite{lin2014microsoft}. The pre-training dataset is either of ImageNet (IN) \cite{5206848}, COCO (CC), or COCO+ (CC+).
}
\footnotesize
\resizebox{1\columnwidth}{!}{
\begin{tabular}{l c c c c c}
\toprule
Method & Model / Dataset  & PVOC12 & CC-Th. & CC-St. & Avg. \\
\midrule
\textit{Global features} \\
$\byol$ \cite{grill2020bootstrap} & $\resnet$50 / CC+      & 13.6 & 9.4 &  8.9 & 10.6 \\
$\dino$ \cite{caron2021emerging} & $\vit$-S/16 / CC        & 5.2  & 9.4 &  14.0 & 9.5 \\
\midrule
\textit{Local features} \\
$\orl$ \cite{xie2021unsupervised} & $\resnet$50 / CC+     & 11.9 & 12.0 &  13.7 & 12.5 \\
$\densecl$ \cite{wang2021dense} & $\resnet$50 / IN       & \underline{18.0} & \textbf{19.2} & 16.9 & \underline{18.0} \\
$\soco$ \cite{wei2021aligning} & $\resnet$50 / IN        & 15.1 & 16.3 & 18.9 & 16.8 \\
$\resim$ \cite{xiao2021region} & $\resnet$50 / IN        & 17.1 & 15.9 & 16.6 & 16.5 \\
$\pixpro$ \cite{xie2021propagate} & $\resnet$50 / IN     & 9.5  & 15.2 & 12.4 & 12.4 \\
$\vicregl$ \cite{bardes2022vicregl} & $\resnet$50 / IN   & 13.9 & 11.2 & 16.0 & 13.7 \\
$\mae$ \cite{he2021masked} & $\vit$-S/16 / CC            & 3.3 &  7.5 & 13.6 & 8.1 \\
$\cpsq$ \cite{wang2022cp2} & $\vit$-S/16 / IN         &  9.5 & 12.9 & 13.6 & 12.0 \\
\midrule
\textit{Ours} \\
\rowcolor{cyan!20} $\mname$ & $\vit$-S/16 / CC  & 16.1 & \underline{17.2} & \underline{20.0} & 17.8 \\
\rowcolor{cyan!20} $\mname$ & $\vit$-S/16 / CC+ & \textbf{20.6} & 17.1 & \textbf{21.9} & \textbf{19.9} \\
\rowcolor{cyan!20} $\mname$ & $\vit$-S/16 / IN & 3.8 & 5.4 & 6.6 & 5.3 \\
\bottomrule
\end{tabular}
\label{table:kmeans_segmentation}
}
\end{table}%

\subsection{Ablation study}
\label{ssec:ablation}
We scrutinize the roles played by different components of $\mname$. Unless otherwise stated, $\lambda_{\text{pos}}=3$, $K_{\text{start}}=12$ and the $\texttt{keys}$ tokens are used for the ablations. Rows corresponding to the chosen setting are \hl{highlighted}.

\vspace{1ex} 
\noindent
\textbf{Weight of the positional cues $\lambda_{\text{pos}}$.} The first element that is ablated is the contribution of the positional bias to the overall performance. In Table~\ref{table:lambdapos_segmentation}, we observe that an increased positional bias leads to improved performance on the unsupervised segmentation task, but a slightly worsened one on the linear segmentation task.
\begin{table}[!ht]
\centering
\caption{
\textbf{Ablation: positional cues weight $\lambda_{\text{pos}}$.} We report the mIoU scores for linear and unsupervised segmentation tasks.
}
\footnotesize
\resizebox{1\columnwidth}{!}{
\begin{tabular}{l c c c c c c c c}
\toprule
 & \multicolumn{2}{c}{PVOC12} && \multicolumn{2}{c}{COCO-Things} && \multicolumn{2}{c}{COCO-Stuff} \\
\cmidrule{2-3} \cmidrule{5-6} \cmidrule{8-9}
$\lambda_{\text{pos}}$ & Unsupervised & Linear && Unsupervised & Linear && Unsupervised & Linear \\
\midrule
0  & 3.4 &  52.2 &&  6.8 & 53.8 && 6.3 & 48.9 \\
1  & 4.0 & 55.8 && 6.8 & 56.6 && 8.3 & 50.3 \\ 
2  & 6.9 & \textbf{56.7} && 7.8 & \textbf{58.0} && 13.3 & \textbf{50.4} \\ 
3  & 15.7 & 56.5 && 12.1 & 56.5 && 17.9 & 50.2 \\ 
\rowcolor{cyan!20} 4  & \textbf{16.4} & 55.0 && \textbf{17.0} & 56.9 && \textbf{20.9} & \textbf{50.4} \\ 
$\infty$  & 2.3 & 55.9 &&  5.4 &  57.2 &&  5.5 & 50.1 \\
\bottomrule
\end{tabular}
\label{table:lambdapos_segmentation}
}
\vspace{-4mm}
\end{table}%

\begin{table}[!ht]
\centering
\caption{
\textbf{Ablation: initial number of centroids $K_{\text{start}}$.} We report the mIoU scores for both the linear and unsupervised segmentation downstream tasks.
}
\footnotesize
\resizebox{1\columnwidth}{!}{
\begin{tabular}{l c c c c c c c c}
\toprule
 & \multicolumn{2}{c}{PVOC12} && \multicolumn{2}{c}{COCO-Things} && \multicolumn{2}{c}{COCO-Stuff} \\
\cmidrule{2-3} \cmidrule{5-6} \cmidrule{8-9}
$K_{\text{start}}$ & Unsupervised & Linear && Unsupervised & Linear && Unsupervised & Linear \\
\midrule
4   & 5.3 & 48.0 && 8.3 & 48.7 && 12.6 & 47.5 \\ 
8   & \textbf{15.8} & 54.8 && \textbf{17.6} & 56.5 && \textbf{23.4} & 49.8\\ 
\rowcolor{cyan!20} 12  & 15.7 & 56.5 && 12.1 & 56.5 && 17.9 & 50.2 \\ 
16  & 10.9 & \textbf{56.9} && 8.0 & \textbf{58.0} && 14.2 & \textbf{50.5} \\ 
\bottomrule
\end{tabular}
\label{table:kstart_segmentation}
}
\end{table}%

\noindent
\textbf{Number of initial centroids $K_{\text{start}}$.} As can be seen in Table~\ref{table:kstart_segmentation}, the linear segmentation scores monotonically increase with the number of initial centroids, whereas for unsupervised segmentation, there seems to be a middle ground.

\vspace{1ex} 
\noindent
\textbf{Type of clustering tokens.}
\label{par:ablation_tokens} 
Table~\ref{table:keys_segmentation} shows that the choice of spatial tokens plays a determinant role in the downstream results and that the multi-clustering approach (Sec.~\ref{sssec:multiple_heads}) can yield a significant boost in performance compared to the case when the clustering uses last spatial tokens $\dense^{(B-1)}$ (\texttt{last}).
\begin{table}[!ht]
\centering
\caption{\textbf{Ablation: type of tokens used for the clustering.} We evaluate the impact of using either of the \texttt{keys}, \texttt{values}, or \texttt{queries} tokens of the last transformer block. We report the mIoU scores for both the linear and unsupervised segmentation downstream tasks.
}
\footnotesize
\resizebox{1\columnwidth}{!}{
\begin{tabular}{l c c c c c c c c}
\toprule
 & \multicolumn{2}{c}{PVOC12} && \multicolumn{2}{c}{COCO-Things} && \multicolumn{2}{c}{COCO-Stuff} \\
\cmidrule{2-3} \cmidrule{5-6} \cmidrule{8-9}
Tokens & Unsupervised & Linear && Unsupervised & Linear && Unsupervised & Linear \\
\midrule
\texttt{last}    & 11.2 & 52.6 && 12.7 & 54.6 && 16.0  & 49.1 \\ 
\texttt{queries} & 8.3  & 55.1 && 7.6  & \textbf{56.5} && 11.9 & 49.8\\ 
\texttt{keys}    & 15.7 & \textbf{56.5} && 12.1 & \textbf{56.5} && 17.9 & \textbf{50.2} \\ 
\rowcolor{cyan!20} \texttt{values}  & \textbf{16.5} & 55.2 && \textbf{16.0} & 55.8 && \textbf{21.5} & 49.5 \\ 
\bottomrule
\end{tabular}
\label{table:keys_segmentation}
}
\end{table}%



\section{Conclusion}
\label{sec:conclusion}
We introduced $\mname$; a novel SSL pre-training method for dense downstream tasks. $\mname$ does not resort to using hand-crafted priors \underline{and} the online clustering algorithm generates pseudo labels for both views in a single and united step. As such, the generated segmentation masks are more coherent and avoid encouraging similarity between objects not univocally represented in both views. $\mname$ is thoroughly evaluated on various  downstream tasks and datasets. In spite of being pre-trained on a medium size scene-centric dataset, the proposed learning paradigm is competitive or outperforms existing methods using ImageNet.

\noindent
{\bf Limitation.}
\label{par:limitations}
As is the case with most dense SSL methods, $\mname$ is only implemented and tested with a single model.
\section*{Acknowledgement}
\label{sec:aknowledgement}
This work is supported by the Personalized Health and Related Technologies (PHRT), grant number 2021/344. This project is also partially funded by the European Research Council (ERC) under the European Union’s Horizon 2020 research and innovation program (Grant Agreement No. 101021347). We acknowledge EuroCC Belgium for awarding this project access to the LUMI supercomputer, owned by the EuroHPC Joint Undertaking, hosted by CSC (Finland) and the LUMI consortium.
\clearpage
{\small
\bibliographystyle{ieee_fullname}
\bibliography{references}

\begin{thebibliography}{10}\itemsep=-1pt

\bibitem{asano2019self}
Yuki~Markus Asano, Christian Rupprecht, and Andrea Vedaldi.
\newblock Self-labelling via simultaneous clustering and representation
  learning.
\newblock {\em arXiv preprint arXiv:1911.05371}, 2019.

\bibitem{bardes2022vicregl}
Adrien Bardes, Jean Ponce, and Yann LeCun.
\newblock Vicregl: Self-supervised learning of local visual features.
\newblock {\em arXiv preprint arXiv:2210.01571}, 2022.

\bibitem{caesar2018coco}
Holger Caesar, Jasper Uijlings, and Vittorio Ferrari.
\newblock Coco-stuff: Thing and stuff classes in context.
\newblock In {\em Proceedings of the IEEE conference on computer vision and
  pattern recognition}, pages 1209--1218, 2018.

\bibitem{caron2018deep}
Mathilde Caron, Piotr Bojanowski, Armand Joulin, and Matthijs Douze.
\newblock Deep clustering for unsupervised learning of visual features.
\newblock In {\em Proceedings of the European conference on computer vision
  (ECCV)}, pages 132--149, 2018.

\bibitem{caron2020unsupervised}
Mathilde Caron, Ishan Misra, Julien Mairal, Priya Goyal, Piotr Bojanowski, and
  Armand Joulin.
\newblock Unsupervised learning of visual features by contrasting cluster
  assignments.
\newblock In {\em Thirty-fourth Conference on Neural Information Processing
  Systems (NeurIPS)}, 2020.

\bibitem{caron2021emerging}
Mathilde Caron, Hugo Touvron, Ishan Misra, Herv{\'e} J{\'e}gou, Julien Mairal,
  Piotr Bojanowski, and Armand Joulin.
\newblock Emerging properties in self-supervised vision transformers.
\newblock {\em arXiv preprint arXiv:2104.14294}, 2021.

\bibitem{chen2020simple}
Ting Chen, Simon Kornblith, Mohammad Norouzi, and Geoffrey Hinton.
\newblock A simple framework for contrastive learning of visual
  representations.
\newblock In {\em International conference on machine learning}, pages
  1597--1607. PMLR, 2020.

\bibitem{cho2021picie}
Jang~Hyun Cho, Utkarsh Mall, Kavita Bala, and Bharath Hariharan.
\newblock Picie: Unsupervised semantic segmentation using invariance and
  equivariance in clustering.
\newblock In {\em Proceedings of the IEEE/CVF Conference on Computer Vision and
  Pattern Recognition}, pages 16794--16804, 2021.

\bibitem{mmseg2020}
MMSegmentation Contributors.
\newblock {MMSegmentation}: Openmmlab semantic segmentation toolbox and
  benchmark.
\newblock \url{https://github.com/open-mmlab/mmsegmentation}, 2020.

\bibitem{cuturi2013sinkhorn}
Marco Cuturi.
\newblock Sinkhorn distances: Lightspeed computation of optimal transport.
\newblock {\em Advances in neural information processing systems},
  26:2292--2300, 2013.

\bibitem{5206848}
Jia Deng, Wei Dong, Richard Socher, Li-Jia Li, Kai Li, and Li Fei-Fei.
\newblock Imagenet: A large-scale hierarchical image database.
\newblock In {\em 2009 IEEE Conference on Computer Vision and Pattern
  Recognition}, pages 248--255, 2009.

\bibitem{dosovitskiy2020image}
Alexey Dosovitskiy, Lucas Beyer, Alexander Kolesnikov, Dirk Weissenborn,
  Xiaohua Zhai, Thomas Unterthiner, Mostafa Dehghani, Matthias Minderer, Georg
  Heigold, Sylvain Gelly, et~al.
\newblock An image is worth 16x16 words: Transformers for image recognition at
  scale.
\newblock {\em arXiv preprint arXiv:2010.11929}, 2020.

\bibitem{pascal-voc-2012}
M. Everingham, L. Van~Gool, C.~K.~I. Williams, J. Winn, and A. Zisserman.
\newblock The {PASCAL} {V}isual {O}bject {C}lasses {C}hallenge 2012 {(VOC2012)}
  {R}esults.
\newblock
  http://www.pascal-network.org/challenges/VOC/voc2012/workshop/index.html.

\bibitem{gidaris2018unsupervised}
Spyros Gidaris, Praveer Singh, and Nikos Komodakis.
\newblock Unsupervised representation learning by predicting image rotations.
\newblock {\em arXiv preprint arXiv:1803.07728}, 2018.

\bibitem{grill2020bootstrap}
Jean-Bastien Grill, Florian Strub, Florent Altch{\'e}, Corentin Tallec,
  Pierre~H Richemond, Elena Buchatskaya, Carl Doersch, Bernardo~Avila Pires,
  Zhaohan~Daniel Guo, Mohammad~Gheshlaghi Azar, et~al.
\newblock Bootstrap your own latent: A new approach to self-supervised
  learning.
\newblock {\em arXiv preprint arXiv:2006.07733}, 2020.

\bibitem{hamilton2022unsupervised}
Mark Hamilton, Zhoutong Zhang, Bharath Hariharan, Noah Snavely, and William~T
  Freeman.
\newblock Unsupervised semantic segmentation by distilling feature
  correspondences.
\newblock {\em arXiv preprint arXiv:2203.08414}, 2022.

\bibitem{hariharan2011semantic}
Bharath Hariharan, Pablo Arbel{\'a}ez, Lubomir Bourdev, Subhransu Maji, and
  Jitendra Malik.
\newblock Semantic contours from inverse detectors.
\newblock In {\em 2011 international conference on computer vision}, pages
  991--998. IEEE, 2011.

\bibitem{he2021masked}
Kaiming He, Xinlei Chen, Saining Xie, Yanghao Li, Piotr Doll{\'a}r, and Ross
  Girshick.
\newblock Masked autoencoders are scalable vision learners.
\newblock {\em arXiv preprint arXiv:2111.06377}, 2021.

\bibitem{he2020momentum}
Kaiming He, Haoqi Fan, Yuxin Wu, Saining Xie, and Ross Girshick.
\newblock Momentum contrast for unsupervised visual representation learning.
\newblock In {\em Proceedings of the IEEE/CVF Conference on Computer Vision and
  Pattern Recognition}, pages 9729--9738, 2020.

\bibitem{henaff2021efficient}
Olivier~J H{\'e}naff, Skanda Koppula, Jean-Baptiste Alayrac, Aaron van~den
  Oord, Oriol Vinyals, and Jo{\~a}o Carreira.
\newblock Efficient visual pretraining with contrastive detection.
\newblock {\em arXiv preprint arXiv:2103.10957}, 2021.

\bibitem{henaff2022object}
Olivier~J H{\'e}naff, Skanda Koppula, Evan Shelhamer, Daniel Zoran, Andrew
  Jaegle, Andrew Zisserman, Jo{\~a}o Carreira, and Relja Arandjelovi{\'c}.
\newblock Object discovery and representation networks.
\newblock {\em arXiv preprint arXiv:2203.08777}, 2022.

\bibitem{hjelm2018learning}
R~Devon Hjelm, Alex Fedorov, Samuel Lavoie-Marchildon, Karan Grewal, Phil
  Bachman, Adam Trischler, and Yoshua Bengio.
\newblock Learning deep representations by mutual information estimation and
  maximization.
\newblock {\em arXiv preprint arXiv:1808.06670}, 2018.

\bibitem{kingma2014adam}
Diederik~P Kingma and Jimmy Ba.
\newblock Adam: A method for stochastic optimization.
\newblock {\em arXiv preprint arXiv:1412.6980}, 2014.

\bibitem{kirillov2019panoptic}
Alexander Kirillov, Kaiming He, Ross Girshick, Carsten Rother, and Piotr
  Doll{\'a}r.
\newblock Panoptic segmentation.
\newblock In {\em Proceedings of the IEEE/CVF Conference on Computer Vision and
  Pattern Recognition}, pages 9404--9413, 2019.

\bibitem{kuhn1955hungarian}
Harold~W Kuhn.
\newblock The hungarian method for the assignment problem.
\newblock {\em Naval research logistics quarterly}, 2(1-2):83--97, 1955.

\bibitem{larsson2017colorization}
Gustav Larsson, Michael Maire, and Gregory Shakhnarovich.
\newblock Colorization as a proxy task for visual understanding.
\newblock In {\em Proceedings of the IEEE conference on computer vision and
  pattern recognition}, pages 6874--6883, 2017.

\bibitem{lebailly2022global}
Tim Lebailly and Tinne Tuytelaars.
\newblock Global-local self-distillation for visual representation learning.
\newblock {\em arXiv preprint arXiv:2207.14676}, 2022.

\bibitem{li2021efficient}
Chunyuan Li, Jianwei Yang, Pengchuan Zhang, Mei Gao, Bin Xiao, Xiyang Dai, Lu
  Yuan, and Jianfeng Gao.
\newblock Efficient self-supervised vision transformers for representation
  learning.
\newblock {\em arXiv preprint arXiv:2106.09785}, 2021.

\bibitem{lin2014microsoft}
Tsung-Yi Lin, Michael Maire, Serge Belongie, James Hays, Pietro Perona, Deva
  Ramanan, Piotr Doll{\'a}r, and C~Lawrence Zitnick.
\newblock Microsoft coco: Common objects in context.
\newblock In {\em European conference on computer vision}, pages 740--755.
  Springer, 2014.

\bibitem{liu2020self}
Songtao Liu, Zeming Li, and Jian Sun.
\newblock Self-emd: Self-supervised object detection without imagenet.
\newblock {\em arXiv preprint arXiv:2011.13677}, 2020.

\bibitem{mo2021object}
Sangwoo Mo, Hyunwoo Kang, Kihyuk Sohn, Chun-Liang Li, and Jinwoo Shin.
\newblock Object-aware contrastive learning for debiased scene representation.
\newblock In {\em Thirty-Fifth Conference on Neural Information Processing
  Systems}, 2021.

\bibitem{noroozi2016unsupervised}
Mehdi Noroozi and Paolo Favaro.
\newblock Unsupervised learning of visual representations by solving jigsaw
  puzzles.
\newblock In {\em European conference on computer vision}, pages 69--84.
  Springer, 2016.

\bibitem{o2020unsupervised}
Pedro~O O~Pinheiro, Amjad Almahairi, Ryan Benmalek, Florian Golemo, and Aaron~C
  Courville.
\newblock Unsupervised learning of dense visual representations.
\newblock {\em Advances in Neural Information Processing Systems},
  33:4489--4500, 2020.

\bibitem{pont20172017}
Jordi Pont-Tuset, Federico Perazzi, Sergi Caelles, Pablo Arbel{\'a}ez, Alex
  Sorkine-Hornung, and Luc Van~Gool.
\newblock The 2017 davis challenge on video object segmentation.
\newblock {\em arXiv preprint arXiv:1704.00675}, 2017.

\bibitem{purushwalkam2020demystifying}
Senthil Purushwalkam and Abhinav Gupta.
\newblock Demystifying contrastive self-supervised learning: Invariances,
  augmentations and dataset biases.
\newblock {\em Advances in Neural Information Processing Systems},
  33:3407--3418, 2020.

\bibitem{selvaraju2021casting}
Ramprasaath~R Selvaraju, Karan Desai, Justin Johnson, and Nikhil Naik.
\newblock Casting your model: Learning to localize improves self-supervised
  representations.
\newblock In {\em Proceedings of the IEEE/CVF Conference on Computer Vision and
  Pattern Recognition}, pages 11058--11067, 2021.

\bibitem{van2021revisiting}
Wouter Van~Gansbeke, Simon Vandenhende, Stamatios Georgoulis, and Luc~V Gool.
\newblock Revisiting contrastive methods for unsupervised learning of visual
  representations.
\newblock {\em Advances in Neural Information Processing Systems},
  34:16238--16250, 2021.

\bibitem{van2021unsupervised}
Wouter Van~Gansbeke, Simon Vandenhende, Stamatios Georgoulis, and Luc Van~Gool.
\newblock Unsupervised semantic segmentation by contrasting object mask
  proposals.
\newblock In {\em Proceedings of the IEEE/CVF International Conference on
  Computer Vision}, pages 10052--10062, 2021.

\bibitem{wang2022cp2}
Feng Wang, Huiyu Wang, Chen Wei, Alan Yuille, and Wei Shen.
\newblock Cp2: Copy-paste contrastive pretraining for semantic segmentation.
\newblock {\em arXiv preprint arXiv:2203.11709}, 2022.

\bibitem{wang2022freesolo}
Xinlong Wang, Zhiding Yu, Shalini De~Mello, Jan Kautz, Anima Anandkumar,
  Chunhua Shen, and Jose~M Alvarez.
\newblock Freesolo: Learning to segment objects without annotations.
\newblock In {\em Proceedings of the IEEE/CVF Conference on Computer Vision and
  Pattern Recognition}, pages 14176--14186, 2022.

\bibitem{wang2021dense}
Xinlong Wang, Rufeng Zhang, Chunhua Shen, Tao Kong, and Lei Li.
\newblock Dense contrastive learning for self-supervised visual pre-training.
\newblock In {\em Proceedings of the IEEE/CVF Conference on Computer Vision and
  Pattern Recognition}, pages 3024--3033, 2021.

\bibitem{wang2022exploring}
Zhaoqing Wang, Qiang Li, Guoxin Zhang, Pengfei Wan, Wen Zheng, Nannan Wang,
  Mingming Gong, and Tongliang Liu.
\newblock Exploring set similarity for dense self-supervised representation
  learning.
\newblock In {\em Proceedings of the IEEE/CVF Conference on Computer Vision and
  Pattern Recognition}, pages 16590--16599, 2022.

\bibitem{wei2021aligning}
Fangyun Wei, Yue Gao, Zhirong Wu, Han Hu, and Stephen Lin.
\newblock Aligning pretraining for detection via object-level contrastive
  learning.
\newblock {\em Advances in Neural Information Processing Systems},
  34:22682--22694, 2021.

\bibitem{wen2022self}
Xin Wen, Bingchen Zhao, Anlin Zheng, Xiangyu Zhang, and Xiaojuan Qi.
\newblock Self-supervised visual representation learning with semantic
  grouping.
\newblock {\em arXiv preprint arXiv:2205.15288}, 2022.

\bibitem{xiao2021region}
Tete Xiao, Colorado~J Reed, Xiaolong Wang, Kurt Keutzer, and Trevor Darrell.
\newblock Region similarity representation learning.
\newblock In {\em Proceedings of the IEEE/CVF International Conference on
  Computer Vision}, pages 10539--10548, 2021.

\bibitem{xie2021detco}
Enze Xie, Jian Ding, Wenhai Wang, Xiaohang Zhan, Hang Xu, Peize Sun, Zhenguo
  Li, and Ping Luo.
\newblock Detco: Unsupervised contrastive learning for object detection.
\newblock In {\em Proceedings of the IEEE/CVF International Conference on
  Computer Vision}, pages 8392--8401, 2021.

\bibitem{xie2021unsupervised}
Jiahao Xie, Xiaohang Zhan, Ziwei Liu, Yew~Soon Ong, and Chen~Change Loy.
\newblock Unsupervised object-level representation learning from scene images.
\newblock {\em Advances in Neural Information Processing Systems},
  34:28864--28876, 2021.

\bibitem{xie2021propagate}
Zhenda Xie, Yutong Lin, Zheng Zhang, Yue Cao, Stephen Lin, and Han Hu.
\newblock Propagate yourself: Exploring pixel-level consistency for
  unsupervised visual representation learning.
\newblock In {\em Proceedings of the IEEE/CVF Conference on Computer Vision and
  Pattern Recognition}, pages 16684--16693, 2021.

\bibitem{yun2022patch}
Sukmin Yun, Hankook Lee, Jaehyung Kim, and Jinwoo Shin.
\newblock Patch-level representation learning for self-supervised vision
  transformers.
\newblock In {\em Proceedings of the IEEE/CVF Conference on Computer Vision and
  Pattern Recognition}, pages 8354--8363, 2022.

\bibitem{zhou2017scene}
Bolei Zhou, Hang Zhao, Xavier Puig, Sanja Fidler, Adela Barriuso, and Antonio
  Torralba.
\newblock Scene parsing through ade20k dataset.
\newblock In {\em Proceedings of the IEEE conference on computer vision and
  pattern recognition}, pages 633--641, 2017.

\bibitem{ziegler2022self}
Adrian Ziegler and Yuki~M Asano.
\newblock Self-supervised learning of object parts for semantic segmentation.
\newblock {\em arXiv preprint arXiv:2204.13101}, 2022.

\end{thebibliography}
}
\clearpage
\newpage

\appendix
\setcounter{table}{0}
\renewcommand{\thetable}{A\arabic{table}}
\setcounter{figure}{0}
\renewcommand{\thefigure}{A\arabic{figure}}
\section*{Appendix}
This Appendix provides additional details and qualitative results organized as follows. In~\Cref{sec:app_datasets}, we thoroughly describe the datasets used in our paper. Additional details about concurrent methods and comparison are discussed in~\Cref{sec:app_implementation_details}. A detailed description of the evaluation protocols is presented in~\Cref{sec:app_evaluation_protocols}. Additional results on semi-supervised video segmentation are reported in~\Cref{sec:app_video_segmentation}. In~\Cref{sec:app_overhead}, we briefly discuss the efficiency and computational overhead of $\mname$. Finally, we provide examples of the clusters found on the combined views of complex scene images with the proposed online clustering algorithm in~\Cref{sec:app_clustering_results}.

\noindent
\section{Datasets}
\label{sec:app_datasets}


\noindent
{\bf COCO.}
\label{par:coco}
The COCO (Microsoft Common Objects in Context) dataset \cite{lin2014microsoft} consists of scene-centric images spanning 91 stuff categories and 80 objects/things categories. The \texttt{train2017}, \texttt{val2017} and \texttt{test2017} splits incorporate approximately 118k, 5k and 41k images, respectively. Additionally, a set of $\sim$123k unlabeled images, \texttt{unlabeled2017}, can be used in conjunction with the \texttt{train2017} split to obtain the so-called COCO+ training set.


\noindent
{\bf COCO-Things.}
\label{par:coco_things}
The COCO-Things dataset follows the implementation of \cite{ziegler2022self}. This dataset is based on COCO images and the panoptic labels of \cite{kirillov2019panoptic}. More precisely, the instance-level labels are merged, and so are the 80 ``things'' categories, yielding the following 12 super-categories: \texttt{electronic}, \texttt{kitchen}, \texttt{appliance}, \texttt{sports}, \texttt{vehicle}, \texttt{animal}, \texttt{food}, \texttt{furniture}, \texttt{person}, \texttt{accessory}, \texttt{indoor}, and \texttt{outdoor}. As the underlying images are the same as in the COCO dataset, so are the training/validation/test splits.

\noindent
{\bf COCO-Stuff.}
\label{par:coco_stuff}
The COCO-Stuff dataset follows the implementation of \cite{ziegler2022self}. The stuff annotations are those of \cite{caesar2018coco}. As for COCO-Things, the 91 ``stuff'' categories are merged into 15 super-categories: \texttt{water}, \texttt{structural}, \texttt{ceiling}, \texttt{sky}, \texttt{building}, \texttt{furniture-stuff}, \texttt{solid}, \texttt{wall}, \texttt{raw-material}, \texttt{plant}, \texttt{textile}, \texttt{floor}, \texttt{food-stuff}, \texttt{ground} and \texttt{window}. This dataset follows the same training/validation/test splits as in the COCO dataset.


\noindent
{\bf PVOC12.}
\label{par:pvoc12}
The PASCAL VOC12 (PVOC12) dataset \cite{pascal-voc-2012} is a scene-centric dataset. The \texttt{trainaug} split relies on the extra annotations of \cite{hariharan2011semantic} such that 10582 images with pixel-level labels can be used for the training phase as opposed to the 1464 segmentation masks initially available. The validation set encompasses 1449 finely annotated images. The dataset spans 20 object classes (+1 background class): \texttt{person}, \texttt{bird}, \texttt{cat}, \texttt{cow}, \texttt{dog}, \texttt{horse}, \texttt{sheep}, \texttt{aeroplane}, \texttt{bicycle}, \texttt{boat}, \texttt{bus}, \texttt{car}, \texttt{motorbike}, \texttt{train}, \texttt{bottle}, \texttt{chair}, \texttt{dining table}, \texttt{potted plant}, \texttt{sofa}, \texttt{tv/monitor} and \texttt{background}.

\noindent
{\bf ADE20K.}
\label{par:ade20k}
The ADE20K dataset \cite{zhou2017scene} is a scene-centric dataset encompassing more than 20K scene-centric images and pixel-level annotations. The labels span 150 semantic categories, including ``stuff'' categories, \eg \texttt{sky}, \texttt{road}, or \texttt{grass}, and ``thing'' categories, \eg \texttt{person}, \texttt{car}, \etc.

\section{Implementation details}
\label{sec:app_implementation_details}
\noindent
\subsection{Comparison with competing methods}
\label{par:app_baselelines}
To compare $\mname$ on an equal footing with concurrent methods, we evaluate all baselines using our evaluation pipeline, except for the evaluation of $\resnet$50 on the semi-supervised video segmentation task which are taken as is from \cite{yun2022patch}. With our implementation, the results were worse than the ones reported in \cite{yun2022patch} or \cite{henaff2022object}; hence we report their results. Furthermore, for $\byol$ \cite{grill2020bootstrap}\footnote{The checkpoint for $\byol$ is provided and trained by the authors of $\orl$ \cite{xie2021unsupervised}.}, $\orl$ \cite{xie2021unsupervised}, $\densecl$ \cite{wang2021dense}, $\soco$ \cite{wei2021aligning}, $\resim$ \cite{xiao2021region}, $\pixpro$ \cite{xie2021propagate}, $\vicregl$ \cite{bardes2022vicregl} and $\cpsq$ \cite{wang2022cp2}, we use publicly available model checkpoints. The only two exceptions are $\mae$ \cite{he2021masked} and $\dino$ \cite{caron2021emerging} methods. Indeed, no public model checkpoint exists for $\vit$-S/16 pre-trained with $\mae$. Since our implementation builds upon $\dino$, it is important to have $\mname$ and $\dino$ models trained in a similar setting for comparison purposes. 

\noindent
{\bf MAE.}
\label{par:app_mae}
The $\vit$-S/16 is pre-trained under $\mae$ framework on the COCO dataset with the following parameters:
\begin{itemize}
\setlength\itemsep{-0.5em}
    \item \texttt{mask\_ratio}: 0.75
    \item \texttt{weight\_decay}: 0.05
    \item \texttt{base\_lr}: 0.00015
    \item \texttt{min\_lr}: 0.0
    \item \texttt{warmup\_epochs}: 40
    \item \texttt{batch\_size}: 256
    \item \texttt{epochs}: 300
\end{itemize}
We use the following decoder architecture: 
\begin{itemize}
\setlength\itemsep{-0.5em}
    \item \texttt{decoder\_embed\_dim}: 512
    \item \texttt{decoder\_depth}: 8
    \item \texttt{decoder\_num\_heads}: 16
\end{itemize}

\noindent
{\bf DINO.}
\label{par:app_dino}
The $\vit$-S/16 is pre-trained under $\dino$ framework on the COCO dataset with the following parameters\footnote{$\mname$ uses the same setting.}:
\begin{itemize}
\setlength\itemsep{-0.5em}
    \item \texttt{out\_dim}: 65536
    \item \texttt{norm\_last\_layer}: false
    \item \texttt{warmup\_teacher\_temp}: 0.04
    \item \texttt{teacher\_temp}: 0.07
    \item \texttt{warmup\_teacher\_temp\_epochs}: 30
    \item \texttt{use\_fp16}: true
    \item \texttt{weight\_decay}: 0.04
    \item \texttt{weight\_decay\_end}: 0.4
    \item \texttt{clip\_grad}: 0
    \item \texttt{batch\_size}: 256
    \item \texttt{epochs}: 300
    \item \texttt{freeze\_last\_layer}: 1
    \item \texttt{lr}: 0.0005
    \item \texttt{warmup\_epochs}: 10
    \item \texttt{min\_lr}: 1e-05
    \item \texttt{global\_crops\_scale}: [0.25, 1.0]
    \item \texttt{local\_crops\_number}: 0
    \item \texttt{optimizer}: adamw
    \item \texttt{momentum\_teacher}: 0.996
    \item \texttt{use\_bn\_in\_head}: false
    \item \texttt{drop\_path\_rate}: 0.1
\end{itemize}

\section{Evaluation protocols}
\label{sec:app_evaluation_protocols}
For all evaluation protocols and models, the evaluation operates on the frozen features of the backbone. The projection heads, if any, are simply discarded. The output features from \texttt{layer4} of $\resnet$50 are used in all downstream tasks. The resulting features have dimension $d=2048$, whereas the spatial tokens of a $\vit$-S/16 have dimension $d=384$ only.  We concatenate the spatial tokens from the last $n_{b}$ transformer blocks, similar to \cite{caron2021emerging}, to compensate for that difference.

\noindent
{\bf Transfer learning via linear segmentation.}
\label{par:app_linear_segmentation}
Our implementation is based on that of \cite{van2021unsupervised,ziegler2022self}. The input images are re-scaled to $448\times448$ pixels and fed to the frozen model. Following existing works \cite{ziegler2022self}, in the case of $\resnet$50, dilated convolutions are used in the last bottleneck layer such that the resolution of the features is identical for all models. Prior to their processing by the linear layer, the features are up-sampled with bilinear interpolation such that the predictions and the ground-truths masks have the same resolution. Unlike previous works \cite{ziegler2022self,bardes2022vicregl,van2021unsupervised}, we use Adam \cite{kingma2014adam} as an optimizer instead of SGD. Indeed, we observe that this led to significant improvements for \underline{all} baselines, indicating that the reported results were obtained in a sub-optimal regime and hence did not fully reflect the quality of the learned features. We report results on the PVOC12 validation set after training the linear layer on the \texttt{trainaug} split for 45 epochs. For the COCO-Things and COCO-Stuff, the linear layer is first trained for 10 epochs on the training set and subsequently evaluated on the validation set. Regardless of the evaluation dataset and model, we find that a learning rate $\texttt{lr=1e-3}$ works well and that the selected number of epochs is sufficient to reach convergence. Note that contrary to \cite{ziegler2022self}, which randomly samples 10\% of the COCO-Things/-Stuff training images, we use the full set of available images to avoid introducing additional randomness in the results. \par

For the evaluation with ADE20K, we rely on MMSegmentation \cite{mmseg2020} and the \textit{40k iterations schedule}. We set the batch size to 16, and we report for each method the best result after trying learning rates in $\{\texttt{1e-03}, \texttt{8e-04}, \texttt{3e-04}, \texttt{1e-04}, \texttt{8e-05} \}$.

\noindent
{\bf Transfer learning via unsupervised segmentation.}
\label{par:app_kmeans_segmentation}
Our implementation is based on that of \cite{van2021unsupervised,ziegler2022self}. The input images are re-scaled to $448\times448$ pixels and fed to the frozen model. Following existing works \cite{ziegler2022self}, in the case of $\resnet$50, dilated convolutions are used in the last bottleneck layer such that the resolution of the features is identical for all models. Similarly to \cite{ziegler2022self}, the ground-truth segmentation masks and features are down-/up-sampled to have the same resolution ($100\times100$). Consequently, we ran K-Means on the spatial features of all images with as many clusters as there are classes in the dataset. A label is greedily assigned to each cluster with Hungarian matching \cite{kuhn1955hungarian}. We report the mean Intersection over Union (mIoU) score averaged over five seeds. Importantly, \cite{ziegler2022self} observed that better results could be obtained by using a larger number of clusters $K$ than the number of classes in the dataset and hereby having clusters of object-parts instead of objects. Indeed, if this approach provides information on the consistency of the features within object-part clusters, it does not tell anything about the inter-object-parts relationship. For instance, the mIoU scores will reflect the ability of features corresponding to ``car wheels'' to be clustered together and similarly for ``car body'' features, but it won't be impacted by the distance of the two clusters from one another, which is undesirable. We report results on the PVOC12, COCO-Things, and COCO-Stuff validation sets.

\noindent
{\bf Semi-supervised video object segmentation.}
The semi-supervised video object segmentation evaluation follows the implementation of \cite{caron2021emerging,yun2022patch}. We report the mean contour-based accuracy $\mathcal{F}_{m}$, mean region similarity $\mathcal{J}_{m}$ and their average $(\mathcal{J} \& \mathcal{F})_{m}$ on the 30 videos from the validation set of the DAVIS'17 \cite{pont20172017}. The following parameters are used:
\begin{itemize}
\setlength\itemsep{-0.5em}
    \item \texttt{n\_last\_frames}: 7
    \item \texttt{size\_mask\_neighborhood}: 12
    \item \texttt{topk}: 5
\end{itemize}

\section{Semi-supervised video segmentation results}
\label{sec:app_video_segmentation}
\begin{table}[!ht]
\centering
\caption{
\textbf{Semi-supervised video object segmentation task}. The frozen spatial features are evaluated on the video segmentation task by nearest neighbor propagation  DAVIS'17 challenge. The mean region similarity $\mathcal{J}_{m}$, mean contour-based accuracy $\mathcal{F}_{m}$, and their average ($\mathcal{J} \&  \mathcal{F})_{m}$ are reported. $\dagger$ indicates results taken from \cite{yun2022patch}.}
\footnotesize
\begin{tabular}{l c c c c c}
\toprule
Method & Model & Dataset & ($\mathcal{J} \&  \mathcal{F})_{m}$ & $\mathcal{J}_{m}$ & $\mathcal{F}_{m}$ \\
\midrule
\textit{Global features} \\
$\dino$ \cite{caron2021emerging} & $\vit$-S/16 & COCO & 57.1 & 55.3 & 58.9 \\
\midrule
\textit{Local features} \\
$\densecl^{\dagger}$ \cite{wang2021dense} & $\resnet$50 & ImageNet & 50.7 & 52.6 & 48.9  \\
$\resim^{\dagger}$ \cite{xiao2021region} & $\resnet$50 & ImageNet & 49.3 & 51.2 & 47.3 \\
 $\detco^{\dagger}$ \cite{xie2021detco} & $\resnet$50 & ImageNet & 56.7 & \textbf{57.0} & 56.4 \\
 $\odin$ \cite{henaff2022object} & $\resnet$50 & ImageNet & 54.1 & 54.3 & 53.9 \\
$\mae$ \cite{he2021masked} & $\vit$-S/16 & COCO & 48.9 & 47.3 & 50.6  \\
$\cpsq$ \cite{wang2022cp2} & $\vit$-S/16 & ImageNet & 53.7 & 51.3 & 56.1 \\
\midrule
\textit{Ours} \\
\rowcolor{cyan!20} $\mname$ & $\vit$-S/16 & COCO  & \underline{57.4} & 55.7 & \underline{59.1} \\
\rowcolor{cyan!20} $\mname$ & $\vit$-S/16 & COCO+ & \textbf{58.4} & \underline{56.5} & \textbf{60.2} \\
\rowcolor{cyan!20} $\mname$ & $\vit$-S/16 & ImageNet & 44.7 & 43.5 & 45.9 \\
\bottomrule
\end{tabular}
\label{table:video_segmentation}
\end{table}%

 
    




Good results are obtained on the semi-supervised video segmentation (Table~\ref{table:video_segmentation}), indicating the ability of $\mname$ to produce features consistent through time and space.


\section{Computational overhead}
\label{sec:app_overhead}
An important property of $\mname$ is that it generates pseudo-labels/cluster assignments online. Consequently, this step must be efficient. In Table~\ref{table:time_oh}, we verify that the operations inherent to the clustering step amount to less than $10\%$ of the total time of the $\mname$ pipeline.
\begin{table}[ht]
\centering
\caption{
\textbf{The runtime of the main operations in $\mname$} for a batch size of 256 samples distributed over 2 Tesla V100. CrOC-specific operations are \hl{highlighted}.
}
\footnotesize
\begin{tabular}{l c c}
\toprule
operation & absolute time [ms] & relative time [\%] \\
\midrule
$f_{t}(\cdot) + \overbar{h}_{t}(\cdot)$ & 177.8 & 21.1\\
$f_{s}(\cdot) + \overbar{h}_{s}(\cdot)$ & 183.9 & 21.9\\
\rowcolor{cyan!20} $\Q^{*} = \mathcal{C}(\cdot)$ & 67.0 & 8\\
\rowcolor{cyan!20} $h_{t}(\cdot) + h_{s}(\cdot)$ & 4.8 & 0.5\\
\texttt{backprop.} + \texttt{EMA} & 408.0 & 48.5 \\
\bottomrule
\textit{total} & 841.5 & 100\\
\bottomrule
\end{tabular}
\label{table:time_oh}
\end{table}

						



\section{Qualitative results}
\label{sec:app_clustering_results}

\label{par:app_clusters_viz}

\begin{figure*}[ht]
    \centering
    \includegraphics[width=\textwidth]{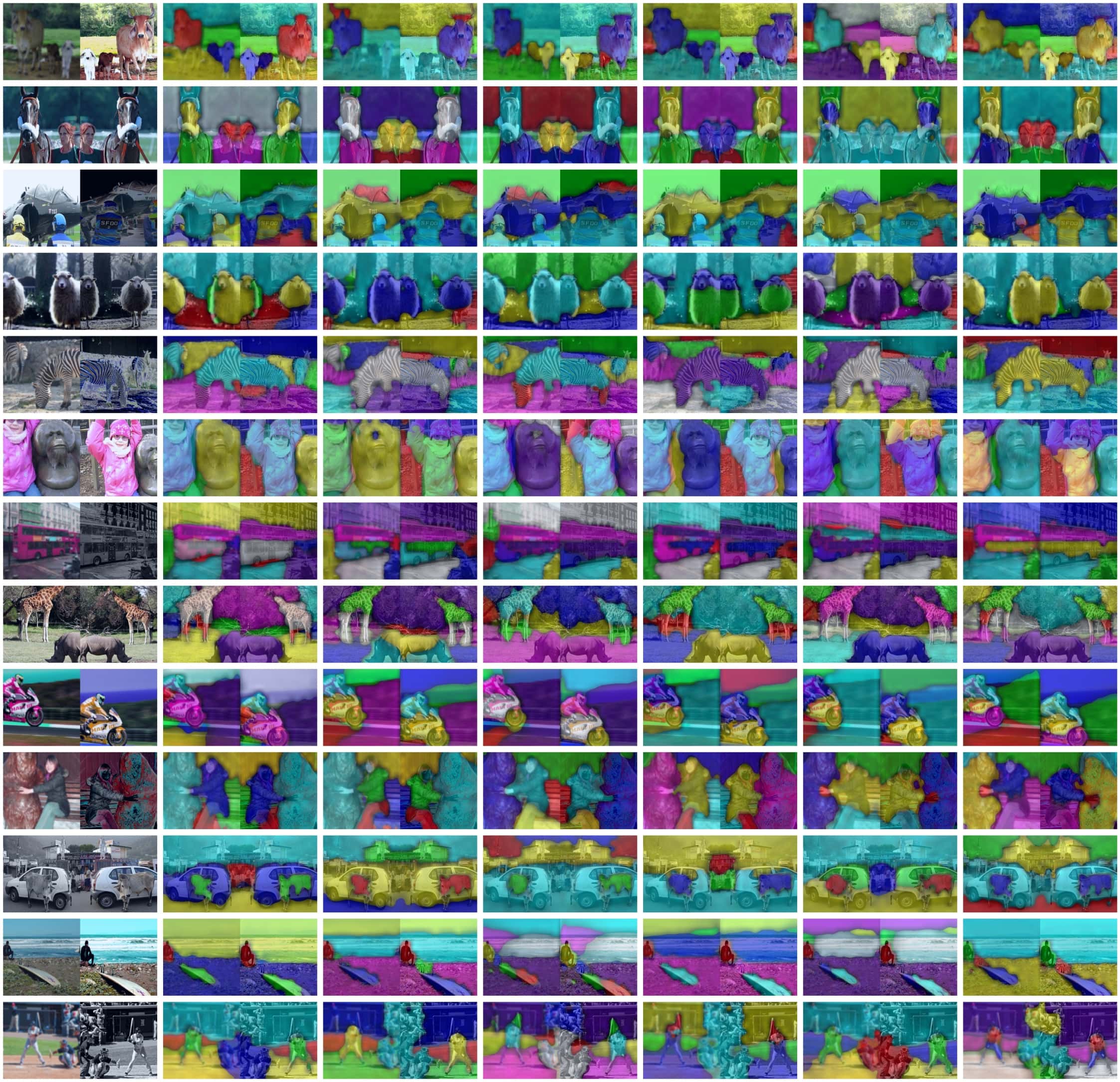}
    \caption{
    \textbf{Illustration of the clusters found online in the space of the combined views}. Rows correspond to combined views and columns to heads of the $\vit$. Bicubic interpolation is used to up-sample the assignments $\Q^{*}$ to the same resolution as the images.}
    \label{fig:clusters}
\end{figure*}

The cluster assignments found by $\mname$'s dedicated online clustering algorithm $\mathcal{C}$ over the combined views are depicted in Fig.~\ref{fig:clusters}. The model used to generate the illustrated assignments is pre-trained on the COCO+ for 300 epochs with $\mname$ and the following meta-parameters: $\lambda_{\text{pos}}=4$, $K_{\text{start}}=12$ and $\texttt{values}$ tokens. During training, we use the same augmentations as in DINO \cite{caron2021emerging}; consequently, we visualize the generated masks based on augmented views in the same manner, such that the results depicted in Fig.~\ref{fig:clusters} truly reflect the consistency enforced by $\mname$.

\end{document}